# Formation Control in Multi-Agent Systems Over Packet Dropping Links

S. Seshadhri[1] and R. Ayyagari[2]*

Industrial Software Systems

ABB Corporate Research

Seshadhri.s@in.abb.com

Department of Instrumentation and Control Engineering

National Institute of Technology, Tiruchirappalli

Tiruchirappalli, India

rkalyn@nitt.edu

**Abstract**

Teams of autonomous agents working in coordination achieve greater efficiency and operation capability than agents performing solo-missions. Multi-agent systems have been investigated widely in the recent past owing to their wide applications and advantages. Formation control is one among the problems being investigated in both control and multi-agent systems paradigm. In formation control teams of agents moving together are required to maintain a pre-defined geometric configuration. Formation control problems have application in vehicle control, unmanned air-craft vehicles (UAVs), consensus and formation control of robots, in industrial robots to name a few. In order to maintain formation, agents in a team need to exchange information like relative displacement, velocity etc. These variables that are exchanged among agents in a team for maintaining formation are called coordination variables, and are used to achieve coordination among agents. Hence, there arises a need to transmit these coordination variables among all the agents in the network. One may visualize that any loss in coordination variable can jeopardize the formation. Communication channels are used for information exchange among agents and are the enabling factor of formation control algorithms.

One major challenge in implementation of formation control problems stems from the packet loss that occur in these shared communication channel. In the presence of packet loss the coordination information among agents is lost. Moreover, there is a move to use wireless channels in formation control applications. It has been found in practice that packet losses are more pronounced in wireless channels, than their wired counterparts.

In our analysis, we first show that packet loss may result in loss of rigidity. In turn this causes the entire formation to fail.   Later, we present an estimation based formation control algorithm that is robust to packet loss among agents. The proposed estimation algorithm employs minimal spanning tree algorithm to compute the estimate of the node variables (coordination variables). Consequently, this reduces the communication overhead required for information exchange. Later, using simulation, we verify the data that is to be transmitted for optimal estimation of these variables in the event of a packet loss. Finally, the effectiveness of the proposed algorithm is illustrated using suitable simulation example.

**7.1. Introduction**

Proliferation of communication channels into control loops have enabled plethora of applications that abstractly engaged the attention of control engineers in the past. Moreover, the presence of communication channel has necessitated a paradigm shift in the way control loops have been analyzed in the past. The usual assumptions of synchronous availability of sensor information and actuation are no longer valid in the presence of communication channels. This is primarily due to packet loss and delay associated with communication channels. Traditional control loops assume continuous availability of feedback data. Data losses that occur in the communication channels make this assumption to be invalid. Furthermore, the tools employed for classical control need to be reformulated/propose new tools for control loops integrated with communication channels.

Control loops integrated with communication channels for information exchange are called networked control systems (NCSs). A detailed review of NCSs can be found in ([1-15] and the references therein). Researchers have investigated NCSs in the recent past owing to their advantages and applications [16-25]. NCSs has many advantages like flexibility, modularity, ease of implementation, reduced wiring, reduction in cost, modularity to name a few. It is now possible to embed control and computing capabilities all along a distributed using pervasive communication channels. Furthermore, scale of the control loops has also increased significantly. Applications of NCSs include intelligent vehicle highway systems (IVHS) [43], tele-robotics [22,50], consensus and cooperation control [40], distributed process control [2,10], wireless sensor networks [8] etc. NCSs differ from distributed control systems (DCS) in the way control components are coupled. In a DCS the control components are loosely coupled i.e. the control loops exist as a decentralized unit that communicate with the centralized controller. The function of the centralized controller is supervision, monitoring and trending. On the other hand, in a NCSs there is a tight interaction among the control components for regular operation of the control loop. The control components interact locally to achieve a common objective as a team.

Motivated by the above developments control scientists have investigated coordinated control among various computing nodes in the network. This has brought into existence the concept of consensus, coordination, and formation control are widely investigated problems in multi-agent domain. The aforesaid problems are generally classified as coordinated control problems in multi-agent domain. Main advantage of coordinated control is its operational efficiency, and the enhanced functionality that can be obtained from the distributed control loop. Teams of agents acting in coordination can produce greater operating efficiency and functionality than individual agents performing solo missions. As an example consider the case of intelligent vehicle highway system (IVHS), maintaining traffic by treating vehicles as agents that coordinate to maintain a pre-specified formation improves the throughput and reduced congestion. Coordination among vehicle (agents) also helps reduce accidents or in other words improve the safety and through put of the traffic. One may conclude from the above discussion that coordination among teams of agents can be used to improve efficiency and can be used to realize goals that were not realizable earlier with individual agents [13-20].

Formation and coordination control are the two problems being widely investigated by control scientists. Consensus control involves team of agents agreeing on a common metric like distance, velocity, etc. The agreement translates into agents meeting or being closer to each other. Randezvous points are used to indicate the agreement among the agents. Formation control has been widely investigated both in control and multi-agent systems paradigm. Accurate maintenance of a geometric configuration among multiple agents moving as a team promises less expensive, more capable systems that can accomplish objectives which might have been impossible with a single agent. The concept of formation control has been studied extensively in the literature with application to the coordination among robots [21-30], UAVs [31], AUVs [32], satellites [33], aircraft [34], and space craft [35]. There are multiple advantages to using formations of agents. These include cost, feasibility, flexibility, accuracy, robustness and energy efficiency. As an example, one may consider the surveillance problem using aircrafts, where in coordination reduces the time required for completing the operation [36].

Various strategies and approaches have been proposed for formation control. These approaches can be roughly categorized into three broad categories, they are: (i) leader-following, (ii) behavioral, and (iii) virtual leader or virtual structure approaches [36]. In the leader-following approach, one of the agent is designated as the leader while others being designated as followers. The basic idea is that the followers track the position and orientation of the leader with some offset. This approach is also called separation-bearing control (SBC) in autonomous robotic consensus [37,38]. There are various variations of this theme including designating multiple leaders, formation of a chain, and other tree topologies. A detailed review of the other two methods, viz., behavioral and virtual structure, is available in [36]. Our investigation is closely related to the leader based strategy or the SBC in consensus control of robots. One drawback in the proposed approaches is the assumption that all agents in the team are informed i.e., the agents know the orientation and position of the all other agents in the team [36]. This requires that coordination information to be available to all the agents in the team at any given instant.

One major challenge in the implementation in multi-agent systems paradigm is the presence of communication channels that are used for achieving coordination among various agents in the team. As communication channels are usually associated with packet loss, maintaining a formation can become increasingly difficult in the presence of packet loss. This is mainly due to the loss of coordination information among various agents in the network. Furthermore, when wireless communication channels are used for implementing formation control loops it has been found in practice that wireless sensor networks are more pronounced to packet loss than their wired counterparts. Typically, in such applications wireless or radio communication is preferred for implementation issues. It is easy to verify that the presence of communication channel makes the assumption that all agents being informed to be invalid. As one might expect that, in order for the formation control algorithm to work, it should be robust to link failures. Furthermore, packet-dropouts can also result in catastrophic outcomes. As an example consider the intelligent vehicle highway system (IVHS) [39], wherein data-loss may result in collision of vehicles. Thus, robustness to packet loss is an important attribute required in any formation control algorithm. In our investigation, we first show that the formation control problem is intractable in the presence of packet losses. Later, we propose an estimation based formation algorithm and finally, we investigate the data to be transmitted in the event of a packet-dropout to reduce the estimation error covariance.

This chapter is organized into six sections including the introduction. In section 7.2, we show that the formation control problem becomes intractable in the presence of packet loss and proceed to discuss the main theme of this investigation. The estimation based formation control algorithm is presented in section 7.3. Investigations on date to be transmitted in the event of a dropped packet for maintaining formation is discussed in section 7.4. Results are summarized in section 7.5, and conclusions are drawn using the obtained results in section 7.6.

## 7.2. Problem Formulation

Consider a formation $\Im$ of agents that are connected using communication links as in Fig. 7.1 (a). The given formation can be can be conveniently represented as a graph, $G_p = \{N, E, W\}$. Where N is the set of nodes, E is the edge-set and W the weights and is the intra-separation distance among the agents as in Fig. 7.1 (b).

**Definition 1:** An agent is said to undergo a rigid motion along a trajectory, only when the Euclidean distance between the agents in the team remains constant all along the trajectory of the agent.

**Definition 2:** The graph in Fig. 1(b) is said to be rigid, if for all position assignments of the nodes, each and every move of the agent preserves the distance between the positions of any pair of vertices in a graph. This condition may be expressed as in (1).

$$\|x_i - x_j\| = C_{ij} \quad \forall \{i, j\} \in E \tag{1}$$

where $C_{ij}$ is the predetermined distance between $i$ and $j$ in the team or the intra-spacing between the agents and $x_i$ and $x_j$ are the relative positions of the agents $i$ and $j$ in the team. The intra-spacing between the agents is similar to the separation in SBC mentioned in section 1. Stated otherwise rigid motion is the only kind of motion the team can undergo. Thus, it is possible to "maintain formation" by keeping the intra-distance spacing constant. This requirement can be given as:

$$\nabla = \|x_i(t) - x_j(t)\| = \|x_i(\tau) - x_j(\tau)\| = C_{ij} \quad \forall [t, \tau] \in \Re^+, \forall \{i, j\} \in E \tag{2}$$

Intuitively rigidity gives the information regarding the minimum number of edges that are needed for maintaining the formation. Rigidity of graphs has been studied for a long time now. One approach to ascertain the rigidity of 2-D planar graph is proposed in [41,42]. We now investigate the conditions for rigidity for the formation control framework. From (1) and (2), we have:

$$\frac{1}{2}\|x_i - x_j\|^2 = C_{ij} \quad \forall \{i,j\} \in E \quad t \geq 0 \tag{3}$$

Assuming smooth trajectory, differentiation of (3) gives

$$\frac{d}{dt}\left(\frac{1}{2}\|x_i - x_j\|^2\right) = (x_i - x_j)^T (\dot{x}_i - \dot{x}_j) \quad \forall \{i,j\} \in E \quad t \geq 0 \tag{4}$$

Little manipulation of (4) leads to

$$R(q)\dot{q} = 0 \tag{5}$$

Where $q = [x_1, x_2, \ldots x_n] \in \Re^{dn}$, n is the cardinality of the vertex set and d is the dimension of the vector. The rigidity matrix given by R(q) is of the order $\Re^{m \times nd}$ with m being the number of edges of the given graph. Given that $q_0$ is a feasible formation, then the graph is generically rigid if and only if [41,42]

$$Rank(R(q_0)) = 3n - 6 \quad if \quad d = 3$$
$$= 2n - 3 \quad if \quad d = 2 \tag{6}$$

One may visualize from (1) and (2) the need for transmitting the relative displacements at any given instant to all the other agents in team. Invariably communication channels are used for transmitting coordination information like relative displacement, velocity etc. Data-loss in communication channels cause the coordination information to be lost. This makes formation intractable as the rigidity of the formation is lost and this can be ascertained from (6). Now, consider the set of points given by

$$d_0 = [d_{01}, d_{02}, \ldots, d_{0n}] \quad d_{0i} \in \Re^{dn} \tag{7}$$

Assuming $d_0$ to meet the rigidity constraints in (6), let us know define the relative error as

$$r_i(t) = x_i(t) - d_{0i} \tag{8}$$

A possible strategy to maintain the formation is to run consensus on (8) above. Assuming the links to be healthy, we have:

$$\dot{r}_i(t) = -\sum_{j \in N_i(t)} (r_i - r_j) \tag{9}$$

Equation (9) illustrates the consensus on (8). It can be verified that

$$\dot{r}_i(t) = \frac{d}{dt}(x_i - d_{0i}) = \dot{x}_i \tag{10}$$

With

$$(r_i - r_j) = (x_i - d_{0i}) - (x_j - d_{0j}) \tag{11}$$

The formation control equation can thus be written as

$$\dot{x}_i = \sum_{j \in N_i(t)} (x_i - x_j) - (d_{0i} - d_{0j}) \tag{12}$$

The main drawback of (12) is that it requires more communication as all the agents should know the position of the other agents in the team. This requires that, all the links to be healthy. In the presence of packet loss the assumption that all agents are informed is invalid. One may conclude from the above discussion that the formation control problem becomes intractable in the presence of packet losses. Thus there is a need to devise an algorithm that is more robust to link failures. In our analysis, we propose an estimation based formation control algorithm among team of multiple –agents connected over a lossy link. We also investigate the data to be transmitted in the event of a packet loss for optimal estimation.

### 7.3. Estimation based formation control algorithm

The first step in the algorithm is to construct a minimum spanning tree (MST) by considering the healthy links in the team. The main requirement to maintain formation from (12) is that the graph $G_p$ should be connected this can be inferred by creating a MST from the healthy links. In our analysis, we employ a greedy algorithm to construct the tree. It is generally seen that as the distance between the agents increases, the packet loss, as well as energy and delays in the channel, also increases. The MST is constructed after leaving out the links with packet dropouts at each time epoch. The algorithm is shown in table 1. The MST with packet-loss between the node 1 and node 2 is shown in Fig. 7.2, and the MST constructed using the algorithm with packet loss between node 5 and node 3 is shown in Fig. 7.3.

The next step in the algorithm is to generate the estimate of the various node variables w.r.t. the leader node. Now consider the graph in Fig. 1(b), w.r.t. one-hop neighbors may then be given as:

$$Z_{21} = x_2 - x_1 + \varepsilon_{21}$$
$$Z_{24} = x_2 - x_4 + \varepsilon_{24}$$

Or more generically as:

$$Z_{ij} = x_i - x_j + \varepsilon_{ij} \tag{13}$$

Equation (13) can be written as

$$Z = Hx_i + \varepsilon \tag{14}$$

Where H is the incidence matrix and it gives the relative displacement of agent $i$ w.r.t. its one-hop neighbors. The least-square estimate of the node variable or the coordination variable can be computed using (15).

$$\hat{x} = (H^T H)^{-1} H^T z \tag{15}$$

Let P be the covariance matrix, then the best linear unbiased estimate (BLUE) is given as

$$\hat{x} = \underbrace{(H^T P^{-1} H)^{-1} H^T P^{-1}}_{L_p} z \tag{16}$$

Where $L_p$ is the graph Laplacian. It is easy to verify that the position of an arbitrary agent at any given time is a linear combination of its own position moves and the position moves of its one-hop neighbors. Thus by considering the reference or leader node (16) can be modified as (18).

$$Z = H_r x_r(k) + H_b x_b(k) + \varepsilon$$

Or
$$Z - H_r x_r(k) = H_b x_b(k) + \varepsilon \tag{17}$$

The best linear unbiased estimate (BLUE) of the one-hop neighbor nodes can then be estimated as:

$$\hat{x}_b^* = (H_b^T P^{-1} H_b)^{-1} H_b^T P^{-1} (Z - H_r^T x_r) \tag{18}$$

Where $H_b$ represents the partitioned matrix containing agents other than the reference node, and $H_b$ is the partitioned incidence matrix considering the reference node. It is seen that equation (19) is similar to the one obtained in [43].

The next step in the algorithm is to construct the estimate of the node variables with the missing links using (19). Once the estimate is available to reference or leader node formation can be maintained using (12). It is easy to visualize from (9) that the position of the agent in the formation depends on its own displacement and that of its one-hop neighbors.

### 7.4. Data to be transmitted in the event of a lost packet for optimal estimation

Two strategies have been proposed in the past for dealing with packet loss in NCSs [47, 48]. They are: (i) Transmitting zero and (ii) Transmitting previous value of the control input in the event of a lost packet. These strategies are called "to zero" and "to hold" respectively in [48]. It has been reported in [47, 48] none of the strategies above can be claimed to be superior to the other. This necessitates simulation or experimentation to select one of the above strategies. Our simulation studies indicate that transmitting a linear combination of the present measurement alongside the estimate of the sensor measurement in the past instant outperforms the "to hold" and "to zero" strategy investigated in [48]. The above result has also been proved theoretically in [2, 40]. In [2], the above result has been extended to the consensus problem among robotic agents, wherein the network of robotic agents shown in Fig.7. 4 is considered. Now, assume that there is a packet loss in link between the agents 1 and agent 3. The simulation of the positions of the robotic agents is shown in Fig. 7.5.

The estimation based formation control algorithm is shown in table 2.

### 7.5. Results and discussions

Consider the formation alongside the initial conditions shown in Fig. 7.6. Given that there is a link failure between agent 1 and agent 2, we compare the performance of the proposed algorithm with that of the other two strategies widely employed in literature -- (i) to transmit zero, or (ii) to transmit past value of the control input [44,45]. The position of the agent 3 and agent 5, and the estimation error covariance after 2 position moves after 50 iterations by transmitting zero is shown in Fig. 7.7. The position of agent 2 and agent 5 by transmitting the measurement available in the previous instant and the estimation error covariance is shown in Fig. 7.8. It is seen that the method of transmitting either zero or the past measurement is not suitable for maintaining the formation.

The positions of agent 3 and agent 5 after two position moves by using the estimation scheme in (19) is shown in Fig. 7.9. It is seen that the proposed estimation scheme is able to maintain the formation in the presence of packet losses. The position agent 3 and agent 5 with packet dropouts in one-link over 50 iterations is shown in Fig. 7.10. It is seen that the scheme performs well and can maintain formation even in the presence of one-link loss over the entire estimation period. The position of agent 3 and agent 5 after 2 position moves by transmitting a linear combination of the past measurement and present estimate is shown in Fig. 7.11. It may be observed that the estimation error covariance is reduced by using the proposed information scheme.

## 7.6. Conclusions

An estimation based formation control algorithm for agents connected over packet dropping links has been proposed in this investigation. It was also shown that the proposed algorithm is robust to link failures. Simulation results show that transmitting a linear combination of past measurement and present estimate reduces the estimation error covariance. Investigations into the maximum packet loss rate that can guarantee proposed formation control algorithm to maintain formation and dynamic formations wherein nodes get attached and tethered are future extensions of this investigation.

List of figures:

| Figure Number | Title |
|---|---|
| Fig.7. 1. (a) | Representation of agent based system |
| Fig. 7.1. (b) | Graph of the team of agent |
| Fig 7.2 | MST with packet-loss in link between node-1 and node-2 |
| Fig. 7.3 | MST with packet-loss in link between node-5 and node-3 |
| Fig. 7.4 | Team of robotic agents with packet loss in link between agent 1 and agent 3 |
| Fig. 7.5 | Position of agents with packet loss using the proposed methodology |
| Fig. 7.6 | Team of agents with initial formation |
| Fig. 7.7 | Position of agent 3 and agent 5 after two position moves by transmitting zero |
| Fig. 7.8 | Position of agent 3 and agent 5 after 2 position moves by transmitting the measurement available at the previous instant |
| Fig. 7.9 | Position of agent 3 and 5 after 2 position moves using the estimation scheme in (19) |
| Fig.7. 10 | Estimated position of agent 3 and agent 5 after 2 position moves with two-packet dropping link for 50 iterations using the estimation scheme in (19) |
| Fig. 7.11 | Position of agent 3 and agent 5 with two packet dropping links for 50 iterations using the proposed information scheme |

Function MST(n,e,w)
Input:
n: number of agents in the team
e: number of links
w: weight associated with the links
T: Tokens from the nodes
G: Graph of the team in terms of edge-set
Initialize Q (arbitrary graph), N(q)=0 is the number of healthy links
For i=1:e
If T(i)==1
N(q)=N(q)+1
Q: Add edge to Q
elseif T(i)==0
Q: remove the edge from (Q)
end
Return Q,N(q)
Input: a(v) leader node edge with minimum weight in Q
W=[];
Q, N(q)
Initialize the tree Tr
While (Tr<N(q)-1)
a(v) be the set containing u
b(v) be the set containing v
if a(u)!= b(v)
then
add edge (v,u) $\leftarrow Tr$
Merge a(u) and b(v) into one cluster
Return Tr

Table 1: Minimum Spanning Tree (MST) algorithm with packet losses

Input:
MST (n,e,w), G- the graph of the formation
Compute Q1: G-Tr
Q2=G-Q1
Determine $Z_b$
Compute $\hat{x}$ using (19) for Q2
Compute $Z_r$

Table 2: Algorithm for formation control over packet dropping links